\documentclass[twoside]{article}
\pdfoutput=1
\usepackage[accepted]{aistats2014}
\usepackage{natbib}
\usepackage{amsmath,amsfonts,amssymb}
\usepackage{graphicx} 
\usepackage{float} 
\usepackage{pgfplots}
\usepgfplotslibrary{groupplots}
\pgfplotsset{compat=newest}

\pgfplotsset{%
  compat=newest,
  table/col sep=comma,
  mark end/.style={%
    scatter,
    scatter src=x,
    scatter/@pre marker code/.code={%
      \pgfmathtruncatemacro\usemark{%
        (\coordindex==(\numcoords-1))
      }
      \ifnum\usemark=0
      \pgfplotsset{mark=none}
      \fi
    },
    scatter/@post marker code/.code={}
  },
}

\begin{document}

\newcommand{\curlyN}{\mathcal N}
\newcommand{\curlyB}{\mathcal B}
\newcommand{\KL}{\textsc{KL}}
\newcommand{\expect}{\mathbb{E}}
\newcommand{\bigO}{\mathcal O}
\newcommand{\zeroV}{\mathbf 0}
\newcommand\identity{\mathbf I}
\newcommand\given{\,|\,}
\renewcommand\d{\mathrm{d}}
\newcommand\half{\tfrac{1}{2}}

\newcommand\latentS{f}
\newcommand\latentV{\mathbf f}
\newcommand\latentM{\mathbf F}

\newcommand\indS{u}
\newcommand\indV{\mathbf u}
\newcommand\indM{\mathbf U}

\newcommand\altLatentS{g}
\newcommand\altLatentV{\mathbf g}
\newcommand\altLatentM{\mathbf G}

\newcommand\dataS{y}
\newcommand\dataV{\mathbf y}
\newcommand\dataM{\mathbf Y}

\newcommand\stepfn{\theta}
\newcommand\probit{\phi}

\newcommand\IPS{x}
\newcommand\IPV{\mathbf x}
\newcommand\IPM{\mathbf X}

\newcommand\indIPS{z}
\newcommand\indIPV{\mathbf z}
\newcommand\indIPM{\mathbf Z}

\newcommand\kernS{k}
\newcommand\kernV{\mathbf k}
\newcommand\kernM{\mathbf K}

\newcommand\Knn{\kernM_{nn}}
\newcommand\Knm{\kernM_{nm}}
\newcommand\Kmn{\kernM_{mn}}
\newcommand\Kmm{\kernM_{mm}}
\newcommand\Qnn{\mathbf{Q}_{nn}}


\twocolumn[

\aistatstitle{Scalable Variational Gaussian Process Classification}

\aistatsauthor{ James Hensman \And Alex Matthews \And Zoubin Ghahramani}

\aistatsaddress{ University of Sheffield \And University of Cambridge \And University of Cambridge } ]

\begin{abstract}
    Gaussian process classification is a popular method with a number of appealing properties. We show how to scale the model within a variational inducing point framework, outperforming the state of the art on benchmark datasets. Importantly, the variational formulation can be exploited to allow classification in problems with millions of data points, as we demonstrate in experiments.
\end{abstract}

\section{Introduction}
Gaussian processes (GPs) provide priors over functions that can be used for many machine learning tasks. In the regression setting, when the likelihood is Gaussian, inference can be performed in closed-form using linear algebra. When the likelihood is non-Gaussian, such as in GP classification, the posterior and marginal likelihood must be approximated. \citet{kuss2005assessing} and \citet{nickisch2008approximations} provide excellent comparisons of several approximate inference methods for GP classification. More recently, \citet{opper2009variational} and \citet{khan2012fast} considered algorithmic improvements to variational approximations in non-conjugate GP models. 

The computational cost of inference in GPs is $\bigO(N^3)$ in general, where $N$ is the number of data. In the regression setting, there has been much interest in low-rank or {\em sparse} approaches to reduce this computational complexity: \citet{quinonero2005unifying} provides a review. Many of these approximation schemes rely on the use of a series of {\it inducing points}, which can be difficult to select. \citet{titsias2009variational} suggests a variational approach (see section \ref{par:titsias}) which provides an objective function for optimizing these points. This variational idea was extended by \citet{hensman2013gaussian}, who showed how the variational objective could be reformulated with additional parameters to enable stochastic optimization, which allows GPs to be fitted to millions of data. 

Despite much interest in approximate GP classification {\em and} sparse approximations, there has been little overlap of the two. The approximate inference schemes which deal with the non-conjugacy of the likelihood generally scale with $\bigO(N^3)$, as they require factorization of the covariance matrix. Two approaches which have addressed these issues simultaneously are the IVM \citep{lawrence2003ivm} and Generalized FITC \citep{naish2007generalized} which both have shortcomings as we shall discuss. 

It is tempting to think that sparse GP classification is simply a case of combining a low-rank approximation to the covariance with one's preferred non-conjugate approximation, but as we shall show this does not necessarily lead to an effective method: it can be very difficult to place the inducing input points, and scalability of the method is usually restricted by the complexity of matrix-matrix multiplications.

For these reasons, there is a strong case for a non-conjugate sparse GP scheme which provides a variational bound on the marginal likelihood, combining the scalability of the stochastic optimization approach with the ability to optimize the positions of the inducing inputs. Furthermore, a variational approach would allow for integration of the approximation within other GP models such as GP regression networks \citep{wilson2012gprn}, latent variable models \citep{lawrence2004gaussian} and deep GPs \citep{damianou2013deep}, as the variational objective could be optimized as part of such a model without fear of overfitting.

The rest of the paper is arranged as follows. In section 2, we briefly cover some background material and existing work. In section 3, we show how variational approximations to the covariance matrix can be used post-hoc to provide variational bounds with non-Gaussian likelihoods. These approaches are not entirely satisfactory, and so in section 4 we provide a variational approach which does not first approximate the covariance matrix, but delivers a variational bound directly. In section 5 we compare our proposals with the state of the art, as well as demonstrating empirically that our preferred method is applicable to very large datasets through stochastic variational optimization. Section 6 concludes.

\section{Background}
\paragraph{Gaussian Process Classification}

Gaussian process priors provide rich nonparametric models of functions. To
perform classification with this prior, the process is `squashed' through a
sigmoidal inverse-link function, and a Bernoulli likelihood conditions the
data on the transformed function values. See \citet{rasmussen2006gaussian} for a review. 

We denote the binary class observations as $\dataV=\{\dataS_n\}_{n=1}^N$, and
then collect the input data into a design matrix $\IPM = \{\IPV_n\}_{n=1}^N$. We
evaluate the covariance function at all pairs of input vectors to build the
covariance matrix $\Knn$ in the usual way, and arrive at a prior for the values of the
GP function at the input points: $p(\latentV) =\curlyN(\latentV\given\zeroV,\Knn)$. 

We denote the probit inverse link function as $\probit(x) = \int_{-\infty}^x\curlyN(a \given 0,\,1)\d a$ and the Bernoulli distribution $\curlyB(\dataS_n\given\probit(\latentS_n)) = \probit(\latentS_n)^{\dataS_n}(1-\probit(\latentS_n))^{1-\dataS_n}$. The joint distribution of data and latent variables becomes
\begin{equation}
	p(\dataV, \latentV) = \prod_{n=1}^N\curlyB(\dataS_n\given\probit(\latentS_n))\;\curlyN(\latentV\given\zeroV,\Knn)\enspace .
	\label{eq:GPC}
\end{equation}
The main object of interest is the posterior over function values
$p(\latentV\given \dataV)$, which must be approximated. We also require an
approximation to the marginal likelihood $p(\dataV)$ in order to optimize (or
marginalize) parameters of the covariance function.  An assortment of
approximation schemes have been proposed (see
\citet{nickisch2008approximations} for a comparison), but they all require
$\bigO(N^3)$ computation. 

\paragraph{Sparse Gaussian Processes for Regression}
\label{par:titsias}
The computational complexity of any Gaussian process method scales with $\bigO(N^3)$ because of the need to invert the covariance matrix $\kernM$. To reduce the computational complexity, many approximation schemes have been proposed, though most focus on regression tasks, see \citet{quinonero2005unifying} for a review. Here we focus on inducing point methods \citep{snelson2005sparse}, where the latent variables are augmented with additional input-output pairs $\indIPM, \indV$, known as `inducing inputs' and `inducing variables'.

The random variables $\indV$ are points on the function in exactly the same way as $\latentV$, and so the joint distribution can be written
\begin{equation}
	p(\latentV, \indV) = \curlyN\left(\left[\begin{array}{c}\latentV\\\indV\end{array}\right]\Big|\zeroV,\,\left[\begin{array}{cc}\Knn&\Knm \\\Knm^\top&\Kmm\end{array}\right]\right)
\end{equation}
where $\Kmm$ is formed by evaluating the covariance function at all pairs of inducing inputs points $\indIPV_m, \indIPV_{m'}$, and $\Knm$ is formed by evaluating the covariance function across the data input points and inducing inputs points similarly. Using the properties of a multivariate normal distribution, the joint can be re-written as
\begin{align}
	p(\latentV, \indV) &= p(\latentV\given\indV)p(\indV)\\
 &= \curlyN(\latentV\given\Knm\Kmm^{-1}\indV, \Knn-\Qnn)\curlyN(\indV\given\zeroV,\,\Kmm)\nonumber
\end{align}
with $\Qnn = \Knm\Kmm^{-1}\Knm^\top$. The joint distribution now takes the form
\begin{equation}
	p(\dataV, \latentV, \indV) = p(\dataV\given\latentV)p(\latentV\given\indV)p(\indV)\enspace .
\end{equation}
To obtain computationally efficient inference, integration over $\latentV$ is approximated. To obtain the popular FITC method (in the case of Gaussian likelihood), a factorization is enforced: $p(\dataV\given\indV) \approx \prod_{n}p(\dataS_n\given\indV)$. To get a variational approximation, the following inequality is used
\begin{equation}
	\log p(\dataV\given\indV)\geq\expect_{p(\latentV\given\indV)}\left[\log p(\dataV\given \latentV)\right]\triangleq\log\tilde p(\dataV\given\indV)\enspace.
\label{eq:basicInequality}
\end{equation}
Substituting this bound on the conditional into the standard expression $p(\dataV)=\int p(\dataV|\indV)p(\indV)\d\indV$ gives a tractable bound on the marginal likelihood for the Gaussian case \citep{titsias2009variational}:
\begin{equation}
	\begin{split}
	\log p(\dataV) \geq &\log\curlyN(\dataV\given \zeroV,\Knm\Kmm^{-1}\Knm^\top + \sigma^2\identity)\\
	&- \frac{1}{2\sigma^2}\textrm{tr}(\Knn-\Qnn)\enspace,
\end{split}
\label{eq:titsias}
\end{equation}
where $\sigma^2$ is the variance of the Gaussian likelihood term. This bound on the marginal likelihood can then be used as an objective function in optimizing the covariance function parameters as well as the inducing input points $\indIPM$. The bound becomes tight when the inducing points are the data points: $\indIPM\! =\! \IPM$ so $\Knm\!=\!\Kmm\!=\!\Knn$ and \eqref{eq:titsias} becomes equal the true marginal likelihood $\log \curlyN(\dataV\given\zeroV,\,\Knn + \sigma^2\identity)$. 

Computing this bound \eqref{eq:titsias} and its derivatives costs $\bigO(NM^2)$. A more computationally scalable bound can be achieved by introducing additional variational parameters \citep{hensman2013gaussian}. Noting that \eqref{eq:titsias} implies an approximate posterior $\tilde p(\indV\given\dataV)$, we introduce a variational distribution $q(\indV) = \curlyN(\indV\given\mathbf{m}, \mathbf{S})$ to approximate this distribution, and applying a standard variational bound, obtain:
\begin{align}
	\log p(\dataV) \geq & \log \curlyN(\dataV\given \Knm\Kmm^{-1}\mathbf{m},\,\sigma^2\identity)\nonumber\\
&- \frac{1}{2\sigma^2}\textrm{tr}(\Knm\Kmm^{-1}\mathbf{S}\Kmm^{-1}\Kmn)\label{eq:svigp}\\
&- \frac{1}{2\sigma^2}\textrm{tr}(\Knn - \Qnn) - \KL[q(\indV)||p(\indV)]\nonumber\enspace.
\end{align}
This bound has a unique optimum in terms of the variational parameters $\mathbf m,\mathbf S$, at which point it is tight to the original sparse GP bound \eqref{eq:titsias}. The advantage of the representation in \eqref{eq:svigp} is that it can be optimized in a stochastic \citep{hensman2013gaussian} or distributed \citep{dai2014gaussian, gal2015distributed} fashion. 

Of course for the Bernoulli likelihood, the required integrals for \eqref{eq:titsias} and \eqref{eq:svigp} are not tractable, but we will build on them both in subsequent sections to build sparse GP classifiers.

\paragraph{Related work}
The informative vector machine (IVM) \citep{lawrence2003ivm} is the first work to
approach sparse GP classification to our knowledge. The idea is to combine
assumed density filtering with a selection heuristic to pick points from
the data $\IPM$ to act as inducing points $\indIPM$: the inducing variables
$\indV$ are then a subset of the latent function variables $\latentV$.

The IVM offers superior performance to support vector machines
\citep{lawrence2003ivm}, along with a probabilistic interpretation. However 
we might expect better
performance by relaxing the condition that the inducing points be a sub-set of
the data, as is the case for regression \citep{quinonero2005unifying}.

%
%
Subsequent work on sparse GP classification \citep{naish2007generalized}
removed the restriction of selecting $\indIPM$ to be a subset of the data
$\IPM$, and ostensibly improved over the assumed density filtering scheme by
using expectation propagation (EP) for inference.

\citet{naish2007generalized} noted that when using the FITC approximation for a Gaussian likelihood, the equivalent prior (see also \citet{quinonero2005unifying}) is
\begin{equation}
p(\latentV) \approx \curlyN(\latentV \given \zeroV,\,\Qnn + \textrm{diag}(\Knn-\Qnn))\enspace .
\end{equation}
The Generalized FITC method combines this approximate prior with a Bernoulli likelihood and uses EP to approximate the posterior. The form of the prior means that the linear algebra within the EP updates is simplified, and a round of updates costs $\bigO(NM^2)$. EP is nested inside an optimization loop, where the covariance hyper-parameters and inducing inputs are optimized against the EP approximation to the marginal likelihood. Computing the marginal likelihood approximation and gradients costs $\bigO(NM^2)$. 

The generalized FITC method works well in practise, often finding solutions which are as good as or better than the IVM, with fewer inducing inputs points \citep{naish2007generalized}. 

%

\paragraph{Discussion}
Despite performing significantly better than the IVM, GFITC does not satisfy our requirements for a scalable GP classifier. There is no clear way to distribute computation or use stochastic optimization in GFITC: we find that it is limited to a few thousand data. Further, the positions of the inducing input points $\indIPM$ can be optimized against the approximation to the marginal likelihood, but there is no guarantee that this will provide a good solution: indeed, our experiments in section 5 show that this can lead to strange pathologies. 

To obtain the ability to place inducing inputs as \citet{titsias2009variational} and to scale as \citet{hensman2013gaussian}, we desire a bound on the marginal likelihood against which to optimize $\indIPM$. 
In the next section, we attempt to build such a variational approximation in the same fashion as FITC, by first using existing variational methods to approximate the covariance, and then using further variational approximate methods to deal with non-conjugacy.


\section{Two stage approaches}
\label{par:MF}

A straightforward approach to building sparse GP classifiers is to separate the
low-rank approximation to the covariance from the non-Gaussian likelihood. In
other words, simply treat the approximation to the covariance matrix as the
prior, and then select an approximate inference algorithm (e.g. from one of
those compared by \citet{nickisch2008approximations}), and then proceed with approximate
inference, exploiting the form of the approximate covariance where possible for
computation saving. 

This is how the generalized FITC approximation was derived. However, as described above we aim to construct variational approximations.

It's possible to construct a variational approach in this mould by using the
fact that the probit likelihood can be written as a convolution of a unit
Gaussian and a step function. Without modifying our original model, we can
introduce a set of additional latent variables $\altLatentV$ which relate to
the original latent variables $\latentV$ through a unit variance isotropic
Gaussian:
\begin{equation}
	p(\dataV, \latentV, \altLatentV) = \prod_{n=1}^N\curlyB(\dataS_n|\theta(\altLatentS_n))\,\curlyN(\altLatentV|\latentV, \identity)\,\curlyN(\latentV|\zeroV, \kernM)
	\label{eq:twostage}
\end{equation}
where $\stepfn$ is a step-function inverse-link. The original model \eqref{eq:GPC} is recovered by marginalization of the additional latent vector: $\int \prod_{n=1}^N\curlyB(\dataS_n\given\stepfn(\altLatentS_n))\,\curlyN(\altLatentV\given\latentV, \identity)\d\altLatentV = \prod_{n=1}^N\curlyB(\dataS_n\given\probit(\latentS_n))$. 

We can now proceed by using a variational sparse GP bound on $\altLatentV$,
followed by a further variational approximation to deal with the non-Gaussian
likelihood.

\subsection{Sparse mean field approach}\label{section:MF}
Encouraged by the success of a (non-sparse) factorizing approximation made by
\citep{hensman2014tilted}, we couple a variational bound on $p(\altLatentV)$
with a mean-field approximation. Substituting the variational bound for a
Gaussian sparse GP \eqref{eq:titsias} with our augmented model
\eqref{eq:twostage} (where $\altLatentV$ in \eqref{eq:twostage} replaces
$\dataV$ in \eqref{eq:titsias}), we arrive at a bound on the joint
distribution:
 \begin{equation} \begin{split}
        p(\dataV,\, \altLatentV) \geq &\prod_{n=1}^N\curlyB(\dataS_n\given\stepfn(\altLatentS_n))\,\curlyN(\altLatentV\given \zeroV,\Qnn+ \identity)\\
	&\exp\{- \half\textrm{tr}(\Knn-\Qnn)\}\enspace .
	\label{eq:mf_sub}
    \end{split}
\end{equation}
Assuming a factorizing distribution $q(\altLatentV) = \prod_n q(\altLatentS_n)$, we obtain a lower bound on the marginal likelihood in the usual variational way, 
and the optimal form of the approximating distribution is a truncated Gaussian
\begin{equation}
	q^\star(\altLatentS_n) = \curlyB(\dataS_n\given\stepfn(\altLatentS_n))\curlyN(\altLatentS_n\given a_n,\tilde \sigma^2_n)/\gamma_n\enspace,
\end{equation}
where $\tilde \sigma^{-2}_n $ is given by the $n^\textrm{th}$ diagonal element of $[\Qnn + \identity]^{-1}$,  $\gamma_n$ are the required normalizers and $a_n$ are free variational parameters. Some cancellation in the bound leads to a tractable expression:
\begin{align}
\log p(\dataV) \geq & \sum_{n=1}^N\log \gamma_n -\half\log|\Qnn + \identity|\nonumber\\
&  - \half\textrm{tr}([\Qnn+\identity]^{-1}\expect_{q(\altLatentV)}[\altLatentV\altLatentV^\top])\nonumber\\
 &+ \half\sum_{n=1}^N\big\{\log \tilde\sigma^2_n + \frac{\expect_{q(\altLatentS_n)}[(a_n - \altLatentS_n)^2]}{\tilde\sigma^2_n}\big\}\nonumber\\
& -\half\textrm{tr}(\Knn-\Qnn).
\end{align}
All the components of this bound can be computed in maximum $\bigO(NM^2)$ time. The expectations under the factorizing variational distribution (and their gradients) are available in closed form. 

Approximate inference can now proceed by optimizing this bound with respect to the variational parameters $\mathbf{a} = \{a_n\}_{n=1}^N$, alongside the hyper-parameters of the covariance function and the inducing points $\indIPM$. Empirically, we find it useful to optimize the variational parameters on an `inner loop', which costs $\bigO(NM)$ per iteration. Computing the relevant gradients outside this loop costs $\bigO(NM^2)$.

\paragraph{Predictions}
To make a prediction for a new latent function value $\altLatentS_\star$ at a test input point $\IPV_\star$, we would like to marginalize across the variational distribution:
\begin{equation}
p(\altLatentS_\star \given \dataV) \approx \int p(\altLatentS_\star\given \altLatentV)q(\altLatentV)\d\altLatentV\enspace .
\label{eq:mf_pred}
\end{equation}
Since this is in general intractable, we approximate it by Monte Carlo. Since
the approximate posterior $q$ is a factorized series of truncated normal
distributions, it is straight-forward to sample from. Prediction of a single
test point costs $\bigO(N)$. 

Alternatively, we can employ a Gaussian approximation to the posterior as suggested by \citet{nickisch2008approximations}. In our sparse formulation, this results in $p(\altLatentV\given\dataV)\approx\curlyN(\altLatentV\given\boldsymbol{\Sigma}\Kmm^{-1}\Kmn\expect_{q(\altLatentV)}[\altLatentV], \boldsymbol{\Sigma})$, with $\boldsymbol{\Sigma}=\Kmm - \Kmn[\Qnn+\identity]^{-1}\Knm$.  Substituting in to \eqref{eq:mf_pred} results in a computational cost of $\bigO(M^2)$ to predict a single test point.

\subsection{A more scalable method?}
The mean-field method in section \ref{section:MF} is somewhat unsatisfactory since the number of variational parameters scales with $N$. The variational parameters are also dependent on each other, and so application of stochastic optimization or distributed computing is difficult. For the Gaussian likelihood case, \citet{hensman2013gaussian} proposes a bound on $\log p(\dataV)$ \eqref{eq:svigp} which can be optimized effectively with $\bigO(M^2)$ parameters. Perhaps it is possible to obtain a scalable algorithm by substituting this bound for $p(\altLatentV)$ as in the above?

Substituting \eqref{eq:svigp} into \eqref{eq:twostage} (again replacing $\dataV$ with $\altLatentV$) results in a tractable integral to obtain a bound on the marginal likelihood, as \citet{hensman2013gaussian} points out. The result is
\begin{equation*}
\begin{split}
\log p(\dataV) &= \log \int p(\dataV\given\altLatentV)p(\altLatentV)\d\altLatentV\\
&\geq \prod_n\curlyB(\dataS_n\given\probit(\kernV_n^\top\Kmm^{-1}\mathbf{m}))\\
&\phantom{\geq}- \half\textrm{tr}(\Knm\Kmm^{-1}\mathbf{S}\Kmm^{-1}\Kmn)\\
&\phantom{\geq}- \half\textrm{tr}(\Knn - \Qnn) - \KL[q(\indV)||p(\indV)]\enspace,
\end{split}
\end{equation*}
where $\kernV_n^\top$ is the $n^\text{th}$ row of $\Knm$. This bound can be optimized in a stochastic or distributed way \citep{tolvanaen2014gaussian}, but has been found to be less effective than might be expected (Owen Thomas, personal communication). To understand why, consider the case where the inducing points are set to the data points $\indIPM=\IPM$, so that $\indV=\latentV$. The bound reduces to 
\begin{equation}
\begin{split}
\log p(\dataV)
&\geq \prod_n\curlyB(\dataS_n\given\probit(\mathbf{m}_n))\\
&\phantom{\geq}- \half\textrm{tr}(\mathbf{S}) - \KL[q(\latentV)||p(\latentV)]\enspace.
\end{split}
\end{equation}
and the optimum occurs where $\mathbf{S}^{-1} = \Knn^{-1} + \identity$, and $\mathbf{m}$ is the maximum {\it a posteriori} (MAP) point. 

This approximation is reminiscent of the Laplace approximation, which also places the mean of the posterior at the MAP point, and approximates the covariance with $\mathbf{S}^{-1} = \Knn^{-1} + \mathbf{W}$, where $\mathbf{W}$ is the Hessian of the log likelihood evaluated at the MAP point. The Laplace approximation is known to be relatively ineffective for classification \citep{nickisch2008approximations}, so it is no surprise that this variational approximation should be ineffective. From here we abandon this bound: the next section sees the construction of a variational bound which is scalable and (empirically) effective.


\section{A single variational bound }
\label{par:KL}
Here we obtain a bound on the marginal likelihood without introducing the additional latent variables as above, and without making factorizing assumptions. We first return to the bound \eqref{eq:basicInequality} on the conditional used to construct the variational bounds for the Gaussian case:
\begin{equation}
\log p(\dataV \given \indV ) \geq \expect_{p(\latentV \given \indV )} \left[ \log p(\dataV \given \latentV ) \right]
\label{eq:conditional_again}
\end{equation}
which is in general intractable for the non-conjugate case. We nevertheless persist, recalling the standard variational equation
\begin{equation}
\log p(\dataV) \geq \expect_{q(\indV)}\left[\log p(\dataV\given \indV)\right] - \KL\left[ q(\indV) || p(\indV ) \right ]\enspace .
\label{eq:standard_vb}
\end{equation}

Substituting \eqref{eq:conditional_again} into \eqref{eq:standard_vb} results in a (further) bound on the marginal likelihood:
\begin{align}
\log p(\dataV) &\geq \expect_{q(\indV)}\left[ \log p(\dataV \given \indV ) \right ] - \KL\left[ q(\indV) || p(\indV ) \right ] \nonumber\\
& \geq \expect_{q(\indV)}\left[ \expect_{p(\latentV | \indV )} \left[ \log p(\dataV | \latentV ) \right] \right] - \KL\left[ q(\indV) || p(\indV ) \right ]\nonumber \\
&= \expect_{q(\latentV)}\left[ \log p(\dataV \given \latentV ) \right] - \KL\left[ q(\indV) || p(\indV ) \right ] 
\label{eq:newbound}
\end{align}
where we have defined:
\begin{equation}
q(\latentV) := \int p(\latentV \given \indV ) q(\indV ) \mathrm{d}\indV\enspace .
\label{eq:qf}
\end{equation}

Consider the case where $q(\indV) = \mathcal{N}( \indV \given \mathbf{m} , \mathbf{S} )$. This gives the following functional form for $q(\latentV)$:
\begin{equation}
q(\latentV) = \mathcal{N}( \latentV \given \mathbf{A} \mathbf{m} \, , \Knn + \mathbf{A} (\mathbf{S}-\Kmm) \mathbf{A}^\top)
\label{eq:posterior_form}
\end{equation}
with $\mathbf{A} = \Knm\Kmm^{-1}$. 

Since in the classification case the likelihood factors as
\begin{equation}
p(\dataV \given \latentV ) = \prod_{i=1}^{N} p(y_i \given f_i )\enspace,
\end{equation}
we only require the marginals of $q(\latentV)$ in order to compute the expectations in \eqref{eq:newbound}. We are left with some one-dimensional integrals of the log-likelihood, which can be computed by e.g. Gauss-Hermite quadrature:
\begin{align}
\log p(\dataV) \geq \sum_{n=1}^N \expect_{q(\latentS_n)}\left[ \log p(\dataS_n\given \latentS_n)\right] - \KL\left[q(\indV)||p(\indV)\right]\enspace .
\label{eq:klsparse}
\end{align} 

Our algorithm then consists of maximizing the parameters of $q(\indV)$ with respect to this bound on the marginal likelihood using gradient based optimization. To maintain positive-definiteness of $\mathbf S$, we represent it using a lower triangular form $\mathbf{S} = \mathbf{L}\mathbf{L}^\top$, which allows us to perform unconstrained optimization. 


\paragraph{Computations and Scalability}
Computing the \textsc{KL} divergence of the bound \eqref{eq:klsparse} requires $\bigO(M^3)$ computations. Since we expect the number of required inducing points $M$ to be much smaller than the number of data $N$, most of the work will be in computing the expected likelihood terms. To compute the derivatives of these, we use the Gaussian identities made familiar to us by \citet{opper2009variational}:
\begin{equation}
\begin{split}
\frac{\partial}{\partial \mu} \expect_{\curlyN(x|\mu, \sigma^2)}\big[ f(x)\big] &= \expect_{\curlyN(x|\mu, \sigma^2)}\big[\frac{\partial}{\partial x }  f(x)\big]\\
\frac{\partial}{\partial \sigma^2} \expect_{\curlyN(x|\mu, \sigma^2)}\big[ f(x)\big] &= \half\expect_{\curlyN(x|\mu, \sigma^2)}\big[\frac{\partial^2}{\partial x^2 }  f(x)\big] \,.
\end{split}
\end{equation}
We can make use of these by substituting $f$ for $\log p(\dataS_n\given \latentS_n)$ and $\mu, \sigma^2$ for the marginals of $q(\latentV)$ in \eqref{eq:klsparse}. These derivatives also have to be computed by quadrature methods, after which derivatives with respect to $\mathbf{m},\mathbf{L},\indIPM$  and any covariance function parameters requires the application of straight-forward algebra. 

We also have the option to optimize the objective in a distributed fashion due to the ease of parallelizing the simple sum over $N$, or in a stochastic fashion by selecting mini-batches of the data at random as we shall show in the following.

\paragraph{Predictions}
Our approximate posterior is given as $q(\latentV, \indV) = p(\latentV\given\indV)q(\indV)$. To make predictions at a set of test points $\IPM_\star$ for the new latent function values $\latentV_\star$, we substitute our approximate posterior into the standard probabilistic rule:
\begin{align}
p(\latentV_\star\given \dataV) &= \int p(\latentV_\star\given \latentV, \indV)p(\latentV,\indV\given \dataV)\d\latentV\d\indV\\
&\approx \int p(\latentV_\star\given \latentV, \indV) p(\latentV\given \indV)q(\indV)\d\latentV\d\indV\\
&= \int p(\latentV_\star\given\indV)q(\indV)\d\indV
\end{align}
where the last line occurs due to the consistency rules of the GP. The integral is tractable similarly to \eqref{eq:posterior_form}, and we can compute the mean and variance of a test-latent $\latentS_\star$ in $\bigO(M^2)$, from which the distribution of the test label $\dataS_\star$ is easily computed. 

\newlength\figureheight
\newlength\figurewidth
\begin{figure*}
\setlength\figureheight{3.8cm}
\setlength\figurewidth{0.24\textwidth}
\centering\includegraphics{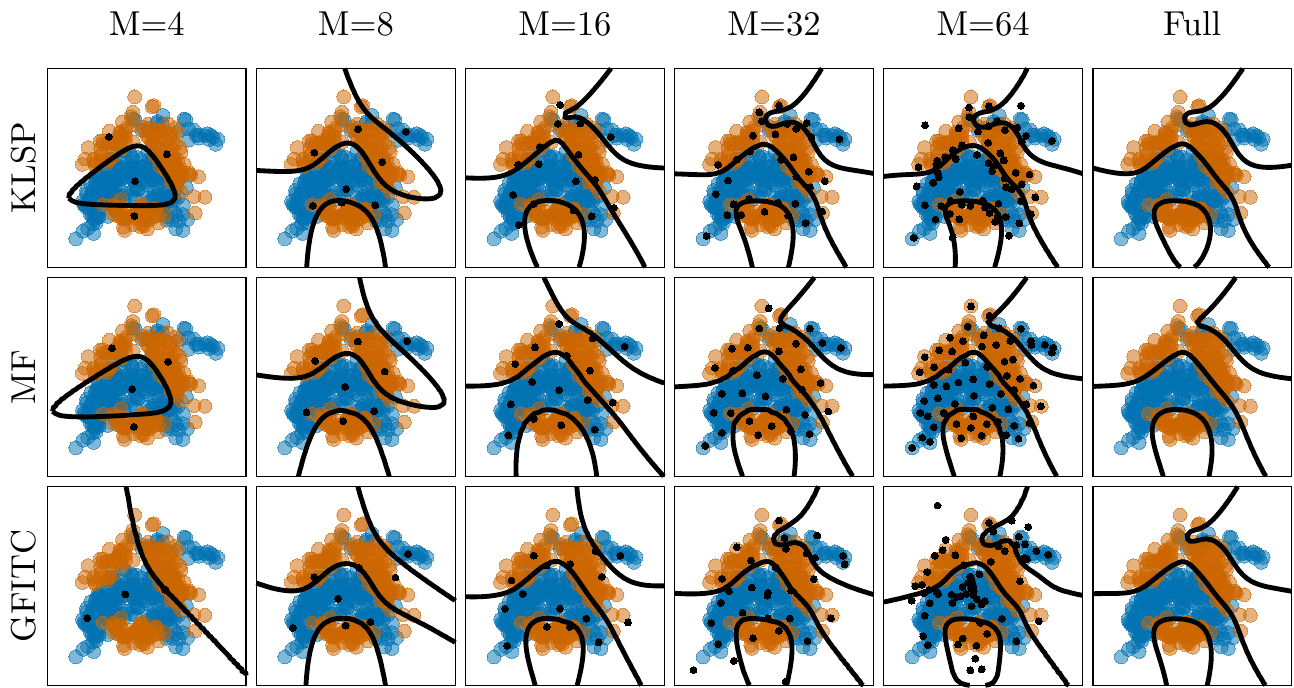}
\vspace{-3mm}
\caption{\label{fig:banana}
The effect of increasing the number of inducing
points for the {\it banana} dataset. Rows represent the KL method, the mean field
method and Generalized FITC, whilst columns show increasing numbers of inducing
points. In each pane, the colored points represent training data, the inducing
inputs are black dots and the decision boundaries are black
lines. The rightmost column shows the result of the equivalent non-sparse
methods}
\end{figure*}

\paragraph{Limiting cases}
To relate this method to existing work, consider two limiting cases of this bound. First, when the inducing variables are equal to the data points $\indIPM\!=\!\IPM$, and second, where the likelihood is replaced with Gaussian noise. 

When $\indIPM\!=\!\IPM$, the approximate posterior in equation \eqref{eq:qf} reduces to $q(\latentV) = \curlyN(\latentV\given{\mathbf{m}},\,\mathbf{S})$. In this special case the number of parameters required to represent the covariance can be reduced to $2N$ \citep{opper2009variational}, and we have recovered the full-Gaussian approximation (the `KL' method described by \citet{nickisch2008approximations}). 

If the likelihood were Gaussian, the expectations in equation \eqref{eq:klsparse} would be computable in closed form, and after a little re-arranging the result is as \citep{hensman2013gaussian}, equation \eqref{eq:svigp}. In this case the bound has a unique solution for $\mathbf{m}$ and $\mathbf{S}$, which recovers the variational bound of \citet{titsias2009variational}, equation \eqref{eq:titsias}. 

In the case where $\indIPM\!=\!\IPM$ and the likelihood is Gaussian, exact inference is recovered.






\section{Experiments}
We have proposed two variational approximations for GP classification. The
first in section \ref{par:MF} comprises a mean-field approximation after making
a variational approximation to the prior over an augmented latent vector. The
second in section \ref{par:KL} proposes to minimize the KL divergence using a
Gaussian approximation at a set of inducing points.  We henceforth refer to
these as the MF (mean-field) and KL methods respectively. 

\paragraph{Increasing the number of inducing points}
To compare the methods with the state-of-the-art Generalized FITC method, we
first turn to the two-dimensional Banana dataset. For all three methods, we
initialized the inducing points using k-means clustering. For the generalized
FITC method we used the implementation provided by
\citet{rasmussen2010gaussian}. For all the methods we used the L-BFGS-B
optimizer \citep{zhu1997algorithm}. 

With the expectation that increasing the number of inducing points should
improve all three methods, we applied $4$ to $64$ inducing points, as shown in
Figure \ref{fig:banana}.  The KL method pulls the inducing points positions
toward the decision boundary, and provides a near-optimal solution with $16$
inducing points. The MF method is less able to adapt the inducing input
positions, but provides good solutions at the same number of inducing points.
The Generalized FITC method appears to pull inducing points towards the
decision boundary, but is unable to make good use of $64$ inducing points,
moving some to the decision boundary and some toward the origin. 

\paragraph{Numerical comparison}
\begin{table*}\caption{\label{table:NumericalComparison} comparison of the performance of our proposed methods and Generalized FITC on benchmark datasets.}
\begin{center}
	{
\begin{tabular}{lllllllll}
       Dataset &              KL &           EP &   \parbox[c]{1.2cm}{EPFitc\\M=8} &  \parbox[c]{1.2cm}{EPFitc\\M=3.0\%} &     \parbox[c]{1.2cm}{KLSp\\M=8} &  \parbox[c]{1.2cm}{KLSp\\M=3.0\%} &     \parbox[c]{1.2cm}{MFSp\\M=8} &  \parbox[c]{1.2cm}{MFSp\\M=3.0\%} \\[2mm]
       \hline \\[-3mm]
       thyroid &    $.11\pm.05$ &  $.11\pm.05$ &  $.15\pm.04$ &     $.13\pm.04$ &  $.13\pm.06$ &   $.09\pm.05$ &  $.22\pm.16$ &   $.15\pm.14$ \\
         heart &    $.46\pm.14$ &  $.43\pm.11$ &  $.42\pm.08$ &     $.44\pm.08$ &  $.42\pm.11$ &   $.47\pm.18$ &  $.41\pm.15$ &   $.43\pm.19$ \\
       twonorm &  $.17\pm.43$ &  $.08\pm.01$ &        Large &     $.08\pm.01$ &  $.08\pm.01$ &   $.09\pm.02$ &        Large &         Large \\
      ringnorm &  $.21\pm.22$ &  $.18\pm.02$ &  $.34\pm.01$ &     $.20\pm.01$ &  $.41\pm.09$ &   $.15\pm.03$ &  $.46\pm.00$ &         Large \\
        german &    $.51\pm.09$ &  $.48\pm.06$ &  $.49\pm.04$ &     $.50\pm.04$ &  $.49\pm.05$ &   $.51\pm.08$ &  $.52\pm.06$ &   $.51\pm.09$ \\
      waveform &  $.23\pm.09$ &  $.23\pm.02$ &  $.22\pm.01$ &     $.22\pm.01$ &  $.23\pm.01$ &   $.25\pm.04$ &  $.28\pm.01$ &         Large \\
        cancer &    $.56\pm.09$ &  $.55\pm.08$ &  $.58\pm.06$ &     $.56\pm.07$ &  $.56\pm.07$ &   $.58\pm.13$ &  $.58\pm.11$ &   $.59\pm.11$ \\
   flare solar &    $.59\pm.02$ &  $.60\pm.02$ &  $.57\pm.02$ &     $.57\pm.02$ &  $.59\pm.02$ &   $.58\pm.02$ &  $.64\pm.05$ &   $.60\pm.04$ \\
      diabetes &    $.48\pm.04$ &  $.48\pm.03$ &  $.50\pm.02$ &     $.51\pm.02$ &  $.47\pm.02$ &   $.51\pm.04$ &        Large &   $.50\pm.04$ \\
\end{tabular}
}
\end{center}

\end{table*}
We compared the performance of the classifiers for a number of commonly used classification data sets. We took ten folds of the data and report the median hold out negative log probability and 2-$\sigma$ confidence intervals. 
For comparison we used the EP FITC implementation from the GPML toolbox \citep{rasmussen2010gaussian} which is generally considered to be amongst the best implementations of this algorithm. 
For the KL and sparse KL methods we found that the optimization behaviour was improved by freezing the kernel hyper parameters at the beginning of optimization and then unfreezing them once a reasonable set of variational parameters has been attained. Table \ref{table:NumericalComparison} shows the result of the experiments. 
In the case where a classifier gives a extremely confident wrong prediction for one or more test points one can obtain a numerically high negative log probability. These cases are denoted as `large'.

The table shows that mean field based methods often give over confident predictions. The results show similar performance between the sparse KL and FITC methods. The confidence intervals of the two methods either overlap or sparse KL is better. As we shall see, Sparse KL runs faster in the timed experiments that follow and can be run on very large datasets using stochastic gradient descent.

\paragraph{Time-performance trade-off}
\begin{figure*}\label{fig:imageNLP}
\setlength\figurewidth{0.9\textwidth}
\setlength\figureheight{7.2cm}
\centering\includegraphics{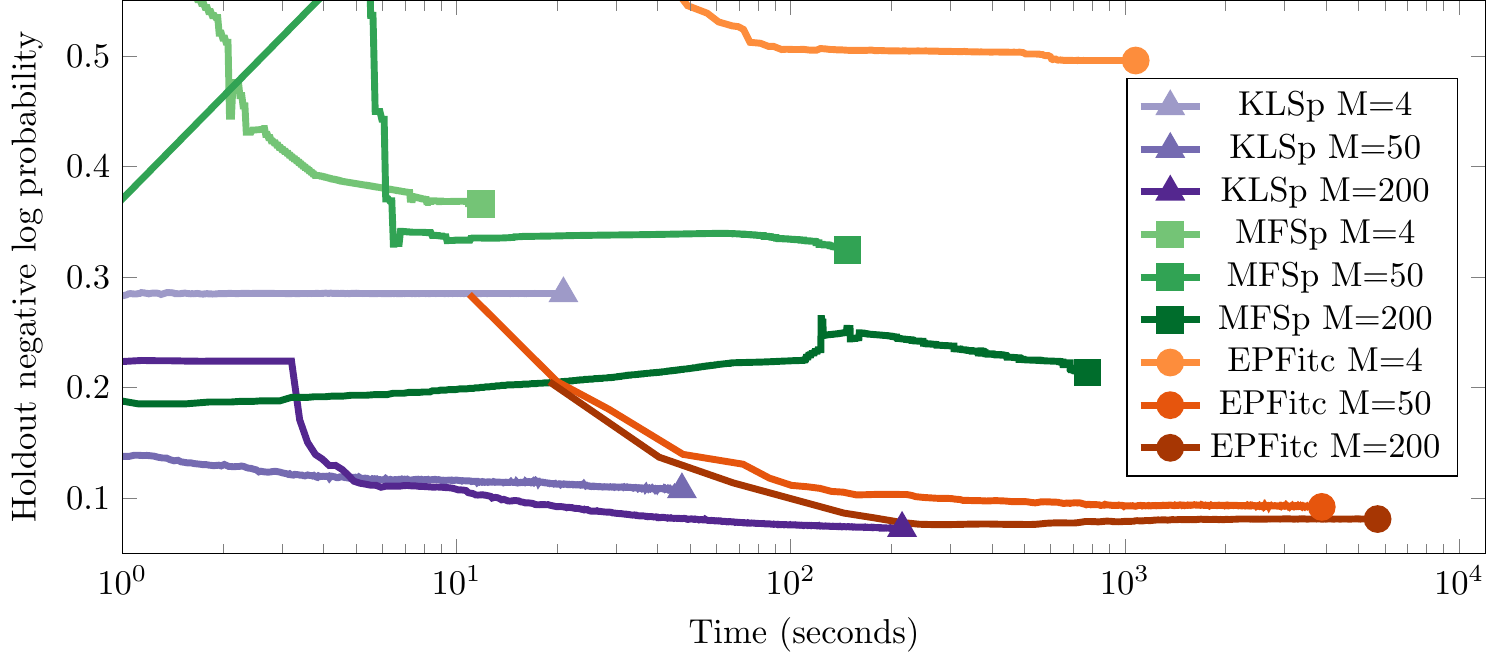}
\vspace{-3mm}
\caption{\label{fig:imageNLP} Temporal performance of the different methods on the {\it image} dataset.}
\end{figure*}

Since all the algorithms used perform optimization there is a trade-off for amount of time spent on optimization against the classification performance achieved. The whole trade-off curve can be used to characterize the efficiency of a given algorithm. 

\citet{naish2007generalized} consider the {\it image} dataset amongst their benchmarks. Of the datasets they consider it is one of the most demanding in terms of the number of inducing points that are required. We ran timed experiments on this dataset recording the wall clock time after each function call and computing the hold out negative log probability of the associated parameters at each step. 

We profiled the MFSp and KLSp algorithms for a variety of inducing point numbers. Although GPML is implemented in MATLAB rather than Python both have access to fast numerical linear algebra libraries. Optimization for the sparse KL and FITC methods was carried out using the LBFGS-B \cite{zhu1997algorithm} algorithm. The mean field optimization surface is challenging numerically and we found that performance was improved by using the scaled conjugate gradients algorithm. We found no significant qualitative difference between the wall clock time and CPU times. 

Figure \ref{fig:imageNLP} shows the results. The \emph{efficient frontier} of the comparison is defined by the algorithm that for any given time achieves the lowest negative log probability. In this case the efficient frontier is occupied by the sparse KL method. Each of the algorithms is showing better performance as the number of inducing points increases. The challenging optimization behaviour of the sparse mean-field algorithm can be seen by the unpredictable changes in hold out probability as the optimization proceeds. The supplementary material shows that in terms of classification error rate this algorithm is much better behaved, suggesting that MFSp finds it difficult to produce well calibrated predictions.

The supplementary material contains plots of the hold out classification accuracy for the {\it image} dataset. It also contains similar plots for the {\it banana} dataset.

\paragraph{Stochastic optimization}
To demonstrate that the proposed KL method can be optimized effectively in a
stochastic fashion, we first turn to the MNIST data set. Selecting the odd digits and even digits to make a binary problem, 
results in a training set with 60000
points. We used ADADELTA method \citep{zeiler2012adadelta} from the {\it climin}
toolbox \citep{climin}. Selecting a step-rate of 0.1, a mini-batch size of 10 with
200 inducing points resulted in a hold-out accuracy of 97.8\%. The mean
negative-log-probability of the testing set was 0.069. It is encouraging that we are able to fit highly nonlinear, high-dimensional decision boundaries with good accuracy, on a dataset size that is out of the range of existing methods. 

Stochastic optimization of our bound allows fitting of GP classifiers to datasets that are larger than previously possible. We downloaded the 
 flight arrival and departure times for every commercial flight in the USA from January 2008 to April 2008. \footnote{Hensman et al. use this data to illustrate regression, here we simply classify whether the delay $\leq$ zero.}
This dataset contains information about 5.9 million flights, including the delay 
in reaching the destination. We build a classifier which was to predict whether any flight was to subject to delay. 

As a benchmark, we first fitted a linear model using scikits-learn \citep{pedregosa2011scikit} which in turns uses LIBLINEAR \citep{fan2008liblinear}. On our randomly selected hold-out set of 100000 points, this achieved a surprisingly low error rate of 37\%, the negative-log probability of the held-out data was 0.642. 

\begin{figure}
\includegraphics{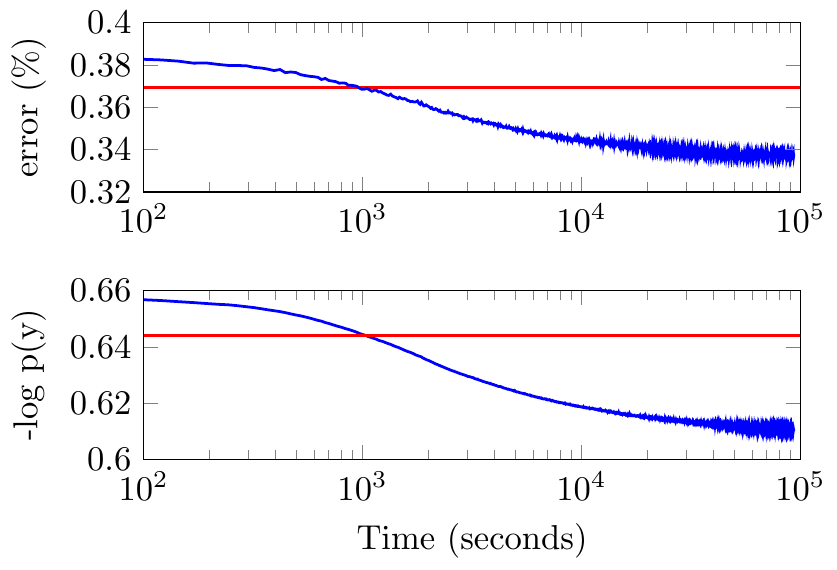}
\vspace{-1cm}
\caption{\label{fig:airline} Performance of the airline delay dataset. The red horizontal line depicts performance of a linear classifier, whilst the blue line shows performance of the stochastically optimized KLsparse method.}
\end{figure}

We built a Gaussian process kernel using the sum of a Matern-$\frac{3}{2}$ and a linear kernel. For each, we introduced parameters which allowed for the scaling of the inputs (sometimes called Automatic Relevance Determination, ARD). Using a similar optimization scheme to the MNIST data above, our method was able to exceed the performance of a linear model in a few minutes, as shown in Figure \ref{fig:airline}. The kernel parameters at convergence suggested that the problem is highly non-linear: the relative variance of the linear kernel was negligible. The optimized lengthscales for the Matern part of the covariance suggested that the most useful features were the time of day and time of year. 

\setlength\figureheight{0.4\linewidth}
\setlength\figurewidth{\linewidth}

\section{Discussion}
We have presented two novel variational bounds for performing sparse GP
classification. The first, like the existing GFITC method, makes an
approximation to the covariance matrix before introducing the non-conjugate
likelihood. These approaches are somewhat unsatisfactory since in performing
approximate inference, we necessarily introduce additional parameters
(variational means and variances, or the parameters of EP factors), which
naturally scale linearly with $N$. 

Our proposed KLSP bound outperforms the state-of-the art GFITC method on benchmark datasets, and is capable of being optimized in a stochastic fashion as we have shown, making GP classification applicable to big data for the first time. 

In future work, we note that this work opens the door for several other GP models: if the likelihood factorizes in $N$, then our method is applicable through Gauss-Hermite quadrature of the log likelihood. 
We also note that it is possible to relax the restriction of $q(\indV)$ to a Gaussian form, and mixture model approximations follow straightforwardly, allowing scalable extensions of \citet{Nguyen2014automated}.

\clearpage

\bibliographystyle{abbrvnat}
\bibliography{bibliography}

\clearpage
\onecolumn\section*{}
\vspace{0.4\textheight}
\begin{center}
	{\large Supplementary Material for :\\}
	\vspace{1em}
	{\Huge Scalable Variational Gaussian Process Classification\\}
	\vspace{1em}
	{\Large anonymous authors}
\end{center}
\renewcommand\thefigure{S.\arabic{figure} }    
\setcounter{figure}{0}

\begin{figure*}
\setlength\figurewidth{\textwidth}
\setlength\figureheight{0.9\textheight}
\centering\includegraphics{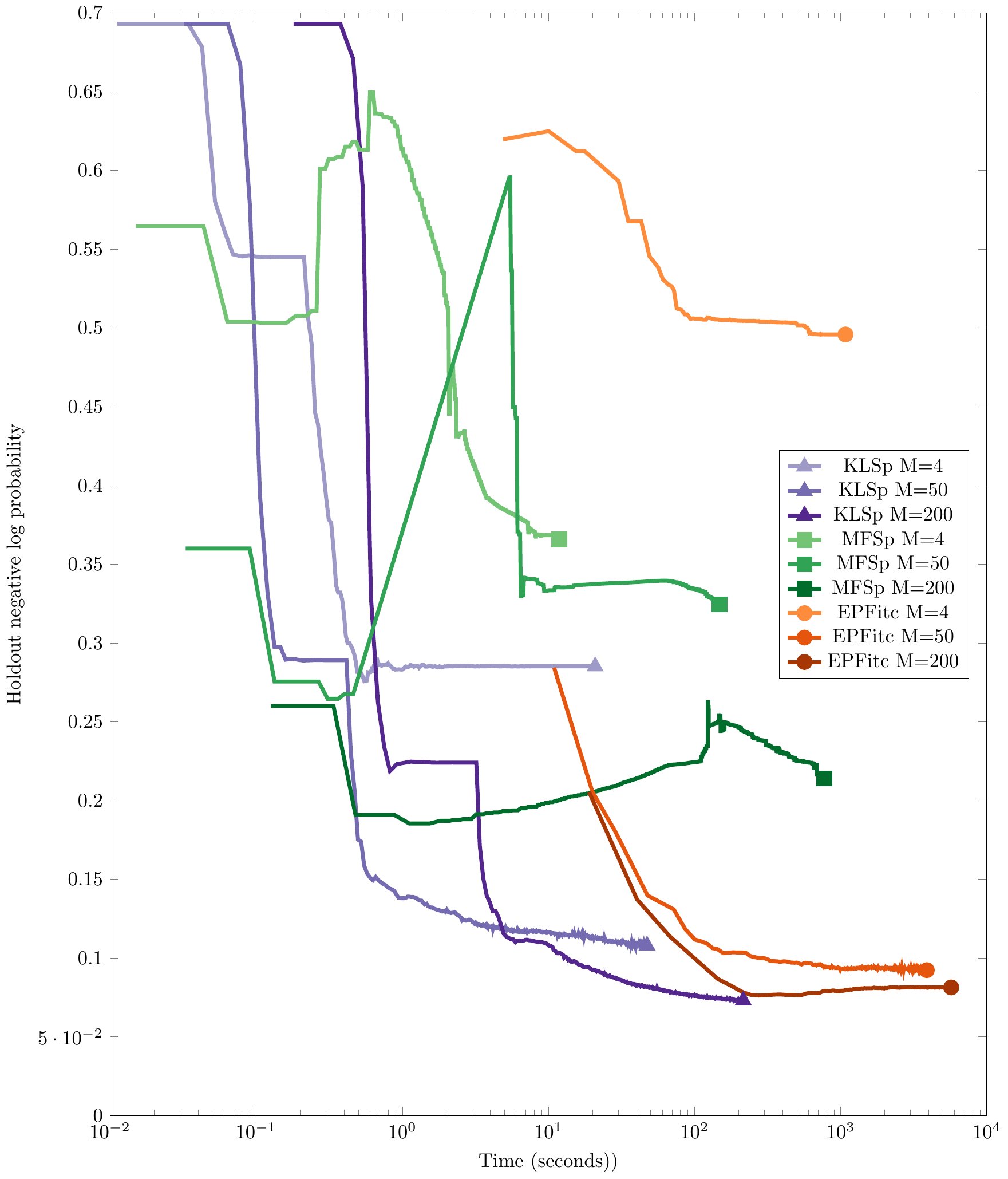}
\caption{\label{fig:image_nlp2} Hold out predictive densities of the different methods on the {\it image} dataset.}
\end{figure*}

\begin{figure*}
\setlength\figurewidth{\textwidth}
\setlength\figureheight{0.9\textheight}
\centering\includegraphics{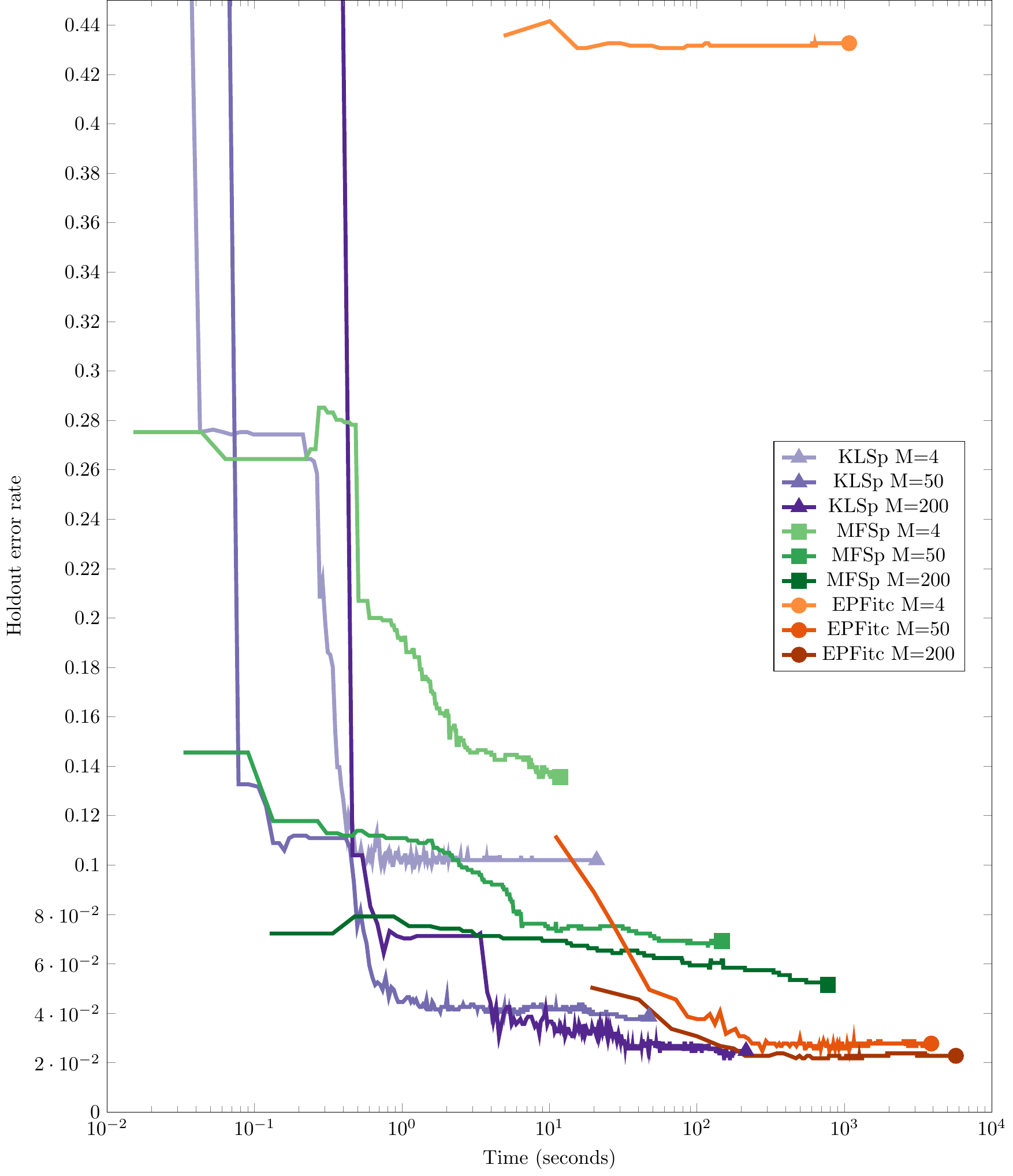}
\caption{\label{fig:image_err} Hold out errors of the different methods on the {\it image} dataset.}
\end{figure*}

\begin{figure*}
\setlength\figurewidth{\textwidth}
\setlength\figureheight{0.9\textheight}
\centering\includegraphics{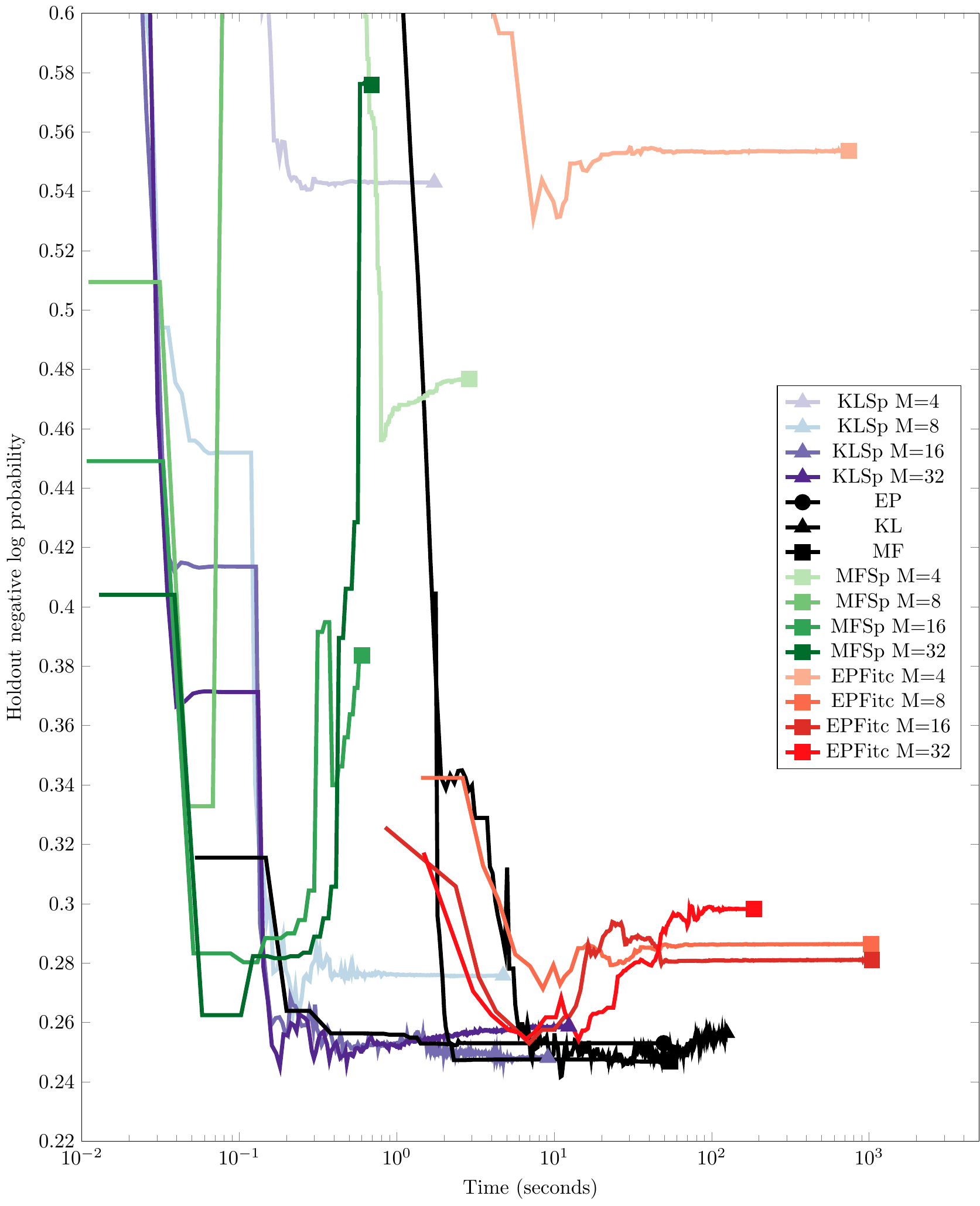}
\caption{\label{fig:banana_nlp} Hold out predictive densities of the different methods on the {\it banana} dataset.}
\end{figure*}

\begin{figure*}
\setlength\figurewidth{\textwidth}
\setlength\figureheight{0.9\textheight}
\centering\includegraphics{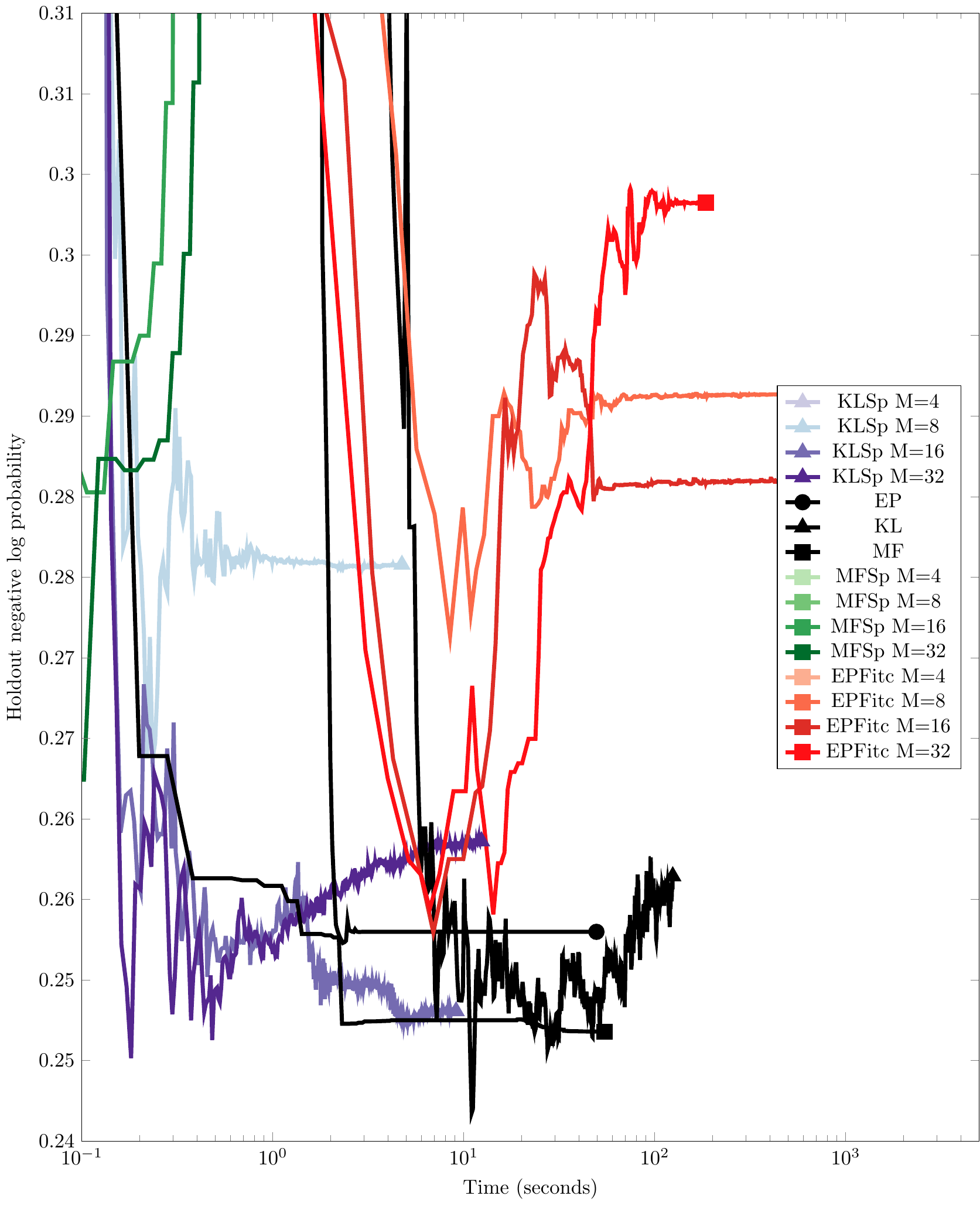}
\caption{\label{fig:banana_nlp_zoom} Hold out predictive densities of the different methods on the {\it banana} dataset.}
\end{figure*}

\begin{figure*}
\setlength\figurewidth{\textwidth}
\setlength\figureheight{0.9\textheight}
\centering\includegraphics{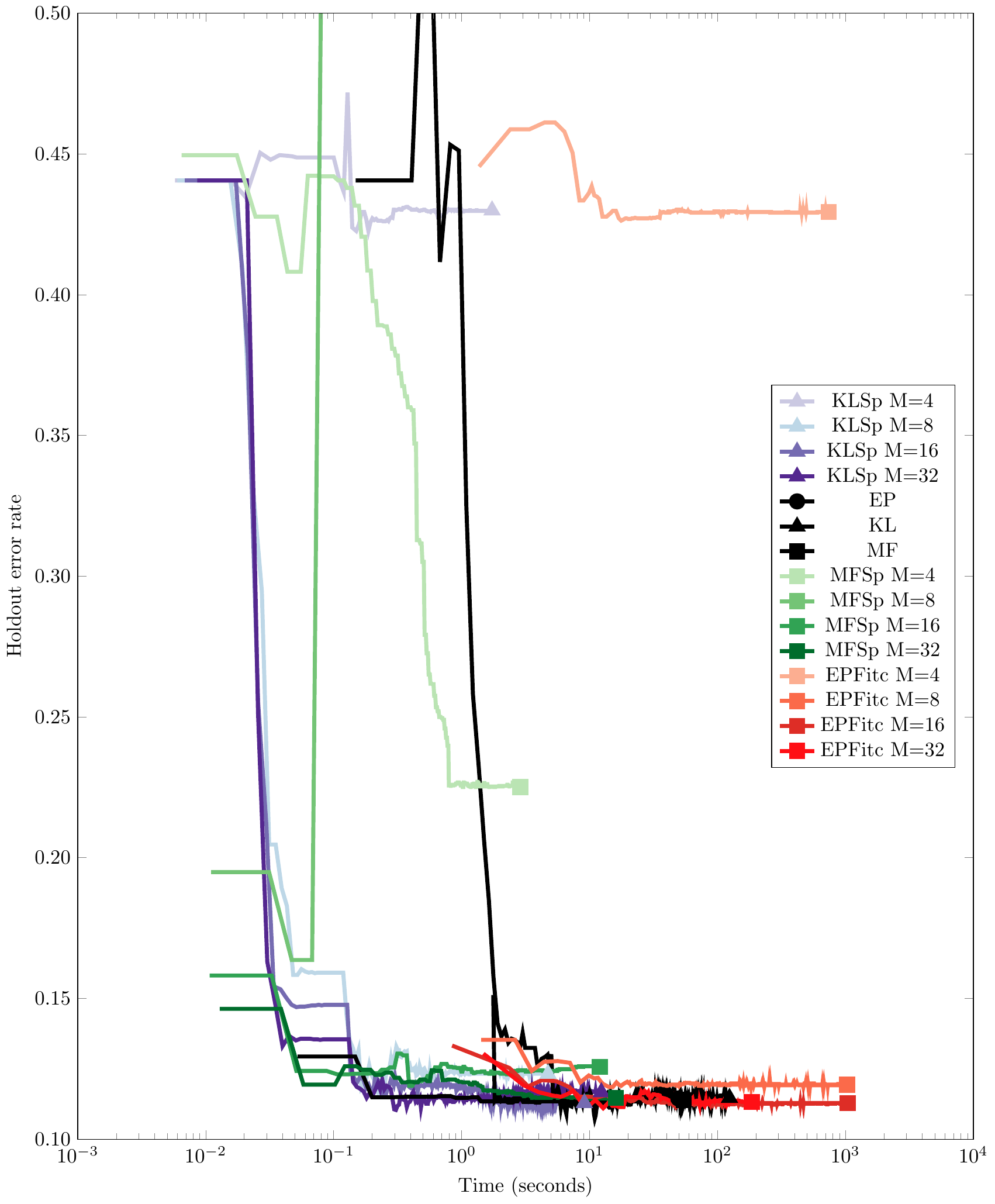}
\caption{\label{fig:banana_err} Hold out errors of the different methods on the {\it banana} dataset.}
\end{figure*}

\begin{figure*}
\setlength\figurewidth{\textwidth}
\setlength\figureheight{0.9\textheight}
\centering\includegraphics{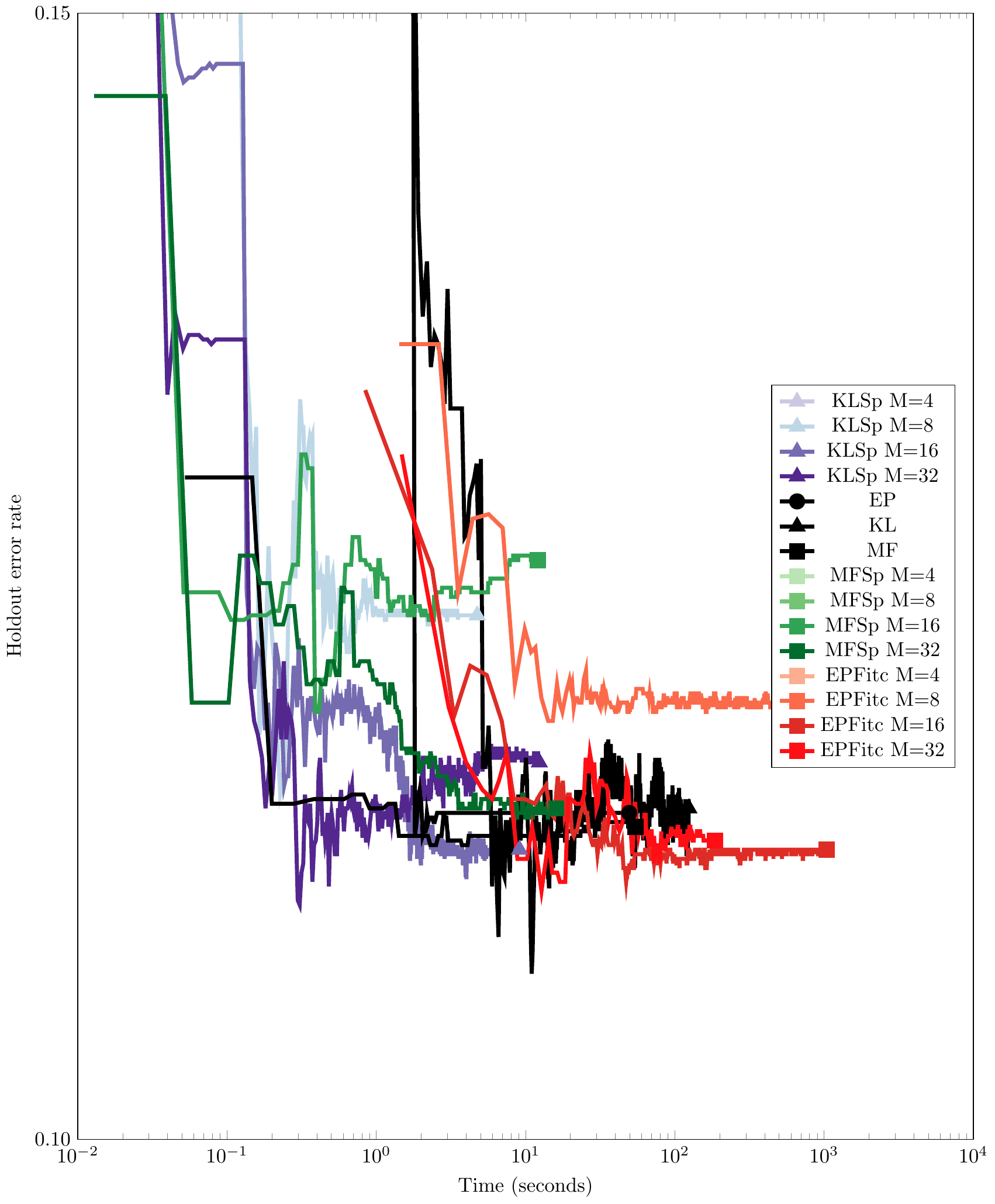}
\caption{\label{fig:banana_err_zoom} Hold out errors of the different methods on the {\it banana} dataset.}
\end{figure*}

\end{document}